\documentclass[journal, onecolumn]{IEEEtran}
\ifCLASSINFOpdf
   \usepackage[pdftex]{graphicx}
   \usepackage[skip=2pt, font=footnotesize, labelsep=period]{caption}
   \usepackage{multirow}
   \usepackage{tabularew}
   \usepackage[flushleft]{threeparttable}
   \usepackage{subfig}
   
\else
\fi
\usepackage[cmex10]{amsmath}
\IEEEoverridecommandlockouts

\makeatletter
\renewcommand{\p@subfigure}{\thefigure(}
\makeatletter
\renewcommand{\figurename}{Figure}
\renewcommand\fnum@figure{\textbf{\figurename\nobreakspace\thefigure}}

\usepackage{etoolbox}
\patchcmd{\section}{\centering}{}{}{}

\renewcommand\thetable{\arabic{table}}
\renewcommand\tablename{Table}
\renewcommand\fnum@table{\textbf{\tablename\nobreakspace\thetable}}

\begin{document}
%
\title{
\begin{flushleft}
Convolutional Spike Timing Dependent Plasticity based Feature Learning in Spiking Neural Networks
\end{flushleft}
}

\author{\begin{flushleft}
\textbf{Priyadarshini Panda\IEEEauthorrefmark{2}\IEEEauthorrefmark{1}, Gopalakrishnan Srinivasan\IEEEauthorrefmark{2}\IEEEauthorrefmark{1} and Kaushik Roy\IEEEauthorrefmark{2}}\\
\IEEEauthorrefmark{2}School of Electrical and Computer Engineering, Purdue University, West Lafayette, US\\
Email:pandap@purdue.edu\\
\IEEEauthorrefmark{1}These authors contributed equally to this work
\end{flushleft}

\vspace{-2.0ex}
}


%


\maketitle
\thispagestyle{plain}
\pagestyle{plain}


%


\vspace{-5.0ex}

\section*{\large\bf{Abstract}}
Brain-inspired learning models attempt to mimic the cortical architecture and computations performed in the neurons and synapses constituting the human brain to achieve its efficiency in cognitive tasks. In this work, we present convolutional spike timing dependent plasticity based feature learning with biologically plausible leaky-integrate-and-fire neurons in Spiking Neural Networks (SNNs). We use shared weight kernels that are trained to encode representative features underlying the input patterns thereby improving the sparsity as well as the robustness of the learning model. We demonstrate that the proposed unsupervised learning methodology learns several visual categories for object recognition with fewer number of examples and outperforms traditional fully-connected SNN architectures while yielding competitive accuracy. Additionally, we observe that the learning model performs out-of-set generalization further making the proposed biologically plausible framework a viable and efficient architecture for future neuromorphic applications. 

\section*{\large\bf{Introduction}}
Mimicking the ``brain'' and its functionalities has been a long-standing dream that has guided researchers in the field of machine learning to devise computational ``machines'' capable of human-like understanding. Consequently, several neuro-inspired artificial models have been proposed \cite{diehl2015unsupervised, lecun2015deep, mnih2015human, ngiam2011multimodal} that perform brain-like functions involving recognition, analytics, and inference among others. In fact, Deep Learning Networks (DLNs), the state-of-the-art machine-learning model, have demonstrated remarkable success across different applications of computer vision even rivalling human performance in certain situations \cite{markoff2011computer,silver2016mastering, he2015delving}. Despite this incredible feat, there still remains a large gap between the human brain and machines that attempt to emulate certain facets of its functionality \cite{cox2014neural, churchland2016computational}.

One such important divide between humans and current machine learning systems is in the size of required training datasets. It has been shown that humans and animals can rapidly learn concepts, often from single training examples. In fact, studies of human concept learning show that humans can accurately learn several visual object categories from minute number of examples \cite{ashby2005human, lake2011one}. This remarkable learning capability of the brain is a direct consequence of the hierarchical organization of the visual cortex and spike-based processing/learning scheme. This, in turn, enables cortical neurons to learn robust object representations and important concepts about a particular input, and further correlate the learnt concepts with other inputs. In contrast, current machine learning approaches require vast quantities of training data to work. In fact, DLNs rely on supervised backpropagation based learning rules that involve millions of trainable parameters or weights, and require millions of labeled data to obtain competitive accuracy on a given application \cite{krizhevsky2012imagenet, han2015learning, iandola2016squeezenet}. 
In this era of `data deluge', labelling every image or input pattern is infeasible. Hence, regardless of their success, the implementation of such large-scale networks has been limited to clouds and servers. It is evident that in order to build intelligent ``machines'' or enable on-device intelligence, it is paramount to use efficient learning models with lesser parameters that can learn real-time with unlabeled data in an unsupervised manner.

Spiking Neural Networks (SNNs), popularly termed as the third generation of neural nets, offer a promising solution for enabling on-chip intelligence. With more biologically realistic perspective on input processing, SNNs perform neural computation by means of spikes in an event-driven fashion. The asynchronous event- or spike-based computations make SNNs an energy-efficient choice over present-day artificial machine learning models that operate with real-valued continuous inputs. Furthermore, SNNs, on account of spike-based processing, can utilize the Spike Timing Dependent Plasticity (STDP) learning mechanism that has been demonstrated to be crucial for memory formation and learning in the brain \cite{masquelier2009competitive, dan2004spike}. STDP enables the SNNs to learn the structure of input patterns without using labels in an unsupervised manner, further making them a suitable alternative for real-time on-chip learning. Consequently, significant efforts have been expended in the recent past to demonstrate the efficacy of SNNs in unsupervised pattern recognition applications \cite{diehl2015unsupervised, querlioz2011simulation, kasabov2013dynamic, masquelier2007unsupervised, kheradpisheh2016stdp, kheradpisheh2016bio}. However, all unsupervised SNN models explored in prior research efforts, though shown to be power-efficient, are based on a fully-connected hierarchical topology \cite{diehl2015unsupervised, querlioz2011simulation} that use the complete training data in a given dataset to get reasonable accuracy. Furthermore, the fully-connected architecture results in a large number of trainable parameters, which in turn increases the training complexity for the SNN. \cite{masquelier2007unsupervised, kheradpisheh2016stdp, kheradpisheh2016bio}, in fact, demonstrate the effectiveness of STDP in learning salient feature representations from images using temporal coding with Integrate-and-Fire (IF) neurons. 

In this work, we present a Convolutional SNN that learns high-level representative features of a given input pattern with lesser connectivity in an unsupervised fashion. The proposed convolutional learning scheme incorporates the shared weight kernel learning of DLNs  crucial for feature extraction. Besides emulating the convolutional hierarchy of the visual cortex, our proposed SNN utilizes the neuro-inspired Leaky-Integrate-and-Fire (LIF) neuronal computational model along with the STDP learning mechanism to conduct synaptic weight updates. The proposed Convolutional SNN is the first of its kind biologically plausible computing model that utilizes spike timing information and inherent latencies to learn the representative features from visual images. Although \cite{masquelier2007unsupervised, kheradpisheh2016stdp, kheradpisheh2016bio} use STDP to learn robust features, the underlying SNN framework or topology in those works consists of several layers of hierachy to perform object recognition. In contrast, our proposed convolutional SNN performs accurate recognition with a single layer topology making it remarkably more efficient than the prior works. The convolutional architecture besides being sparsely connected also enhances the robustness of the SNN based recognition system. The model inherently captures the spatial and rotational variance of the system in natural images. On account of the shared weight kernel learning with STDP, the network discovers general representations of the visual input and associates similarities across different classes of inputs to obtain an improved and robust SNN with lesser complexity as compared to the conventional fully-connected SNN models. 

While the energy-related implications of the proposed design is evident, the proposed SNN, owing to the bio-inspired spike based learning and computing scheme, offers a key advantage of learning from lesser number of training examples while yielding competitive accuracy. This result has a direct correlation with natural intelligence or learning in the brain. In fact, our results on the widely used MNIST digit recognition dataset \cite{lecun1998mnist}, demonstrate that the Convolutional SNN trained on one set of classes (for instance, digits `$6$' and `$7$') is able to identify images from a  completely different class (for instance, digits `$9$' and `$1$'). This result shows that the proposed model can perform out-of-set generalization \cite{cox2014neural}, wherein the system can even recognize classes that it has never encountered in any prior training. This significant result addresses the so-called ‘open-set problem’ \cite{scheirer2014probability, dean2013fast} commonly encountered across all learning models (artificial/spiking).

Next, we discuss the fundamentals of SNNs followed by the proposed Convolutional SNN topology and associated learning methodology. We subsequently detail the simulation framework and present results highlighting the enhanced feature extraction capability and consequent benefits of our Convolutional SNN relative to the conventional fully-connected SNN across multiple benchmark datasets in unsupervised environments.
\section*{\large\bf{Background on Spiking Neural Networks}}
\begin{figure}[!t]
\centering
\includegraphics[width = 6.8in]{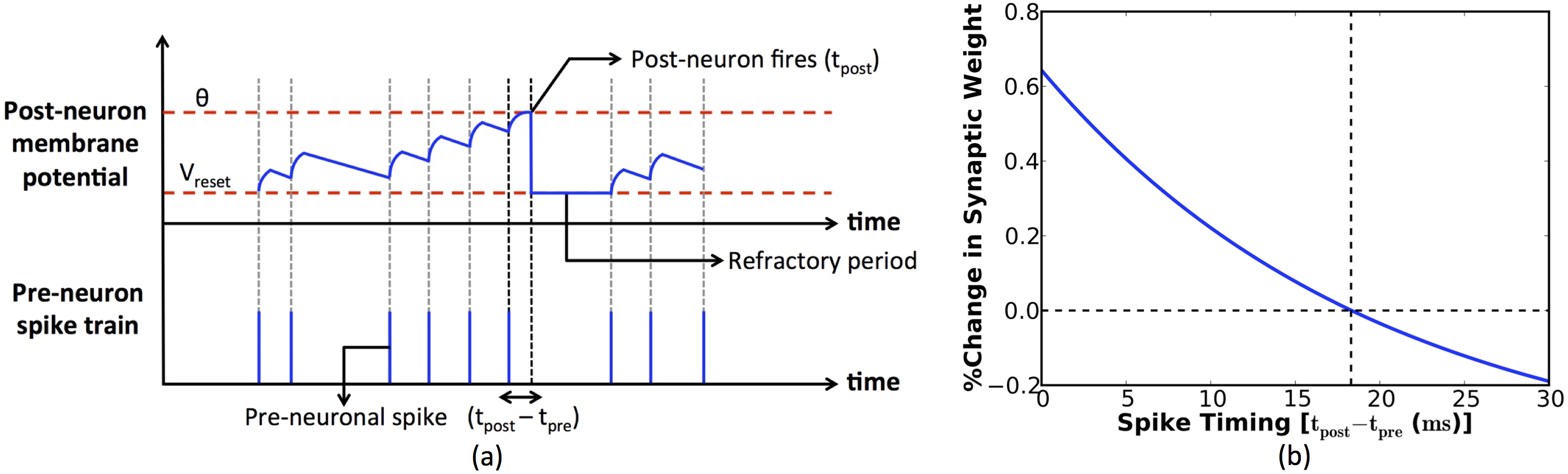}
\caption{\textbf{(a)} Leaky integrate-and-fire dynamics of a spiking neuron. At any given time-step, each input pre-neuronal spike is modulated by the corresponding synaptic weight to generate a resultant current into the post-neuron. This causes an increase in the neuronal membrane potential, which subsequently leaks in the absence of the input spikes. The neuron fires an output spike at a suitable time instant, $t_{post}$, when its membrane potential crosses a specific threshold ($\theta$). The potential is thereafter reset ($V_{reset}$), and the neuron is prevented from spiking for an interval of time known as the refractory period. \textbf{(b)} Change in the synaptic strength based on the temporal correlation in a pair of pre- and post-neuronal spikes ($t_{post}-t_{pre}$) as mandated by the power-law weight-dependent plasticity model formulated in (\ref{eq:STDP}), for a learning rate ($\eta$) of 0.002, time constant ($\tau$) of 20$ms$, offset of 0.4, maximum synaptic weight ($w_{max}$) of unity, previous weight value ($w$) of 0.5, and exponential factor ($\mu$) of 0.9.}
\label{fig:LIF_Dynamics_STDP}
\end{figure}

\subsection*{\normalsize\bf{Computational Model of the Neurons and Synapses}}
Biologically inspired neural computing paradigms typically consist of a layer of input (pre) neurons connected by weighted synapses to the output (post) neurons. The input pre-neuronal (pre-synaptic) spikes are modulated by the synaptic weights to produce a resultant current into the output neurons. We use the Leaky Integrate-and-Fire (LIF) model \cite{diehl2015unsupervised} illustrated in Fig. \ref{fig:LIF_Dynamics_STDP}(a) to emulate the spiking neuronal dynamics. Each post-neuron integrates the current leading to an increase in its membrane potential, which decays exponentially following the removal of the input spikes. The neuron fires an output (post-synaptic) spike, if its membrane potential exceeds a definite threshold. The potential is subsequently reset, and the neuron is restrained from spiking for a certain ensuing duration of time designated as the refractory period. 

\subsection*{\normalsize\bf{Synaptic Plasticity}}
Spike Timing Dependent Plasticity (STDP) is popularly used to achieve unsupervised learning in networks of spiking neurons. STDP postulates that the strength of a synapse is dependent on the degree of temporal correlation in the spiking activity of the connected pre- and post-neuron. In this work, we use the power-law weight-dependent model \cite{diehl2015unsupervised} of synaptic plasticity depicted in Fig. \ref{fig:LIF_Dynamics_STDP}(b). The synaptic weight update is calculated as
\begin{equation} \label{eq:STDP}
\Delta w_{STDP} = \eta \times [exp(\frac{t_{post}-t_{pre}}{\tau}) - offset] \times [w_{max} - w]^\mu
\end{equation}
where $\Delta w_{STDP}$ is the change in the synaptic weight, $\eta$ is the learning rate, $exp(.)$ denotes the exponential function, $t_{pre}$ and $t_{post}$ are respectively the time instant of a pair of pre- and post-synaptic spikes, $\tau$ is the time constant, $w_{max}$ is the maximum constraint imposed on the synaptic weight, and $w$ is the previous weight value. The synaptic strength is increased (potentiated) if a pre-neuron subsequently causes the connected post-neuron to fire and the difference in the respective spiking instants is small, which signifies a strong causal relationship between the corresponding pair of neurons. On the other hand, the synaptic strength is decreased (depressed) for larger spike time differences as determined by the offset in (\ref{eq:STDP}). We further note that the amount of weight update has an exponential dependence (governed by $\mu$ in (\ref{eq:STDP})) on the previous value of synaptic weight. Larger the preceding weight value, smaller is the ensuing weight change and vice versa. The non-linear weight updates ensure gradual increase (decrease) in synaptic strength towards the maximum (minimum) value, which is desirable for efficient learning.

\subsection*{\normalsize\bf{Fully-connected SNN for Pattern Recognition}}
\begin{figure}[!t]
\centering
\includegraphics[width = 5.5in]{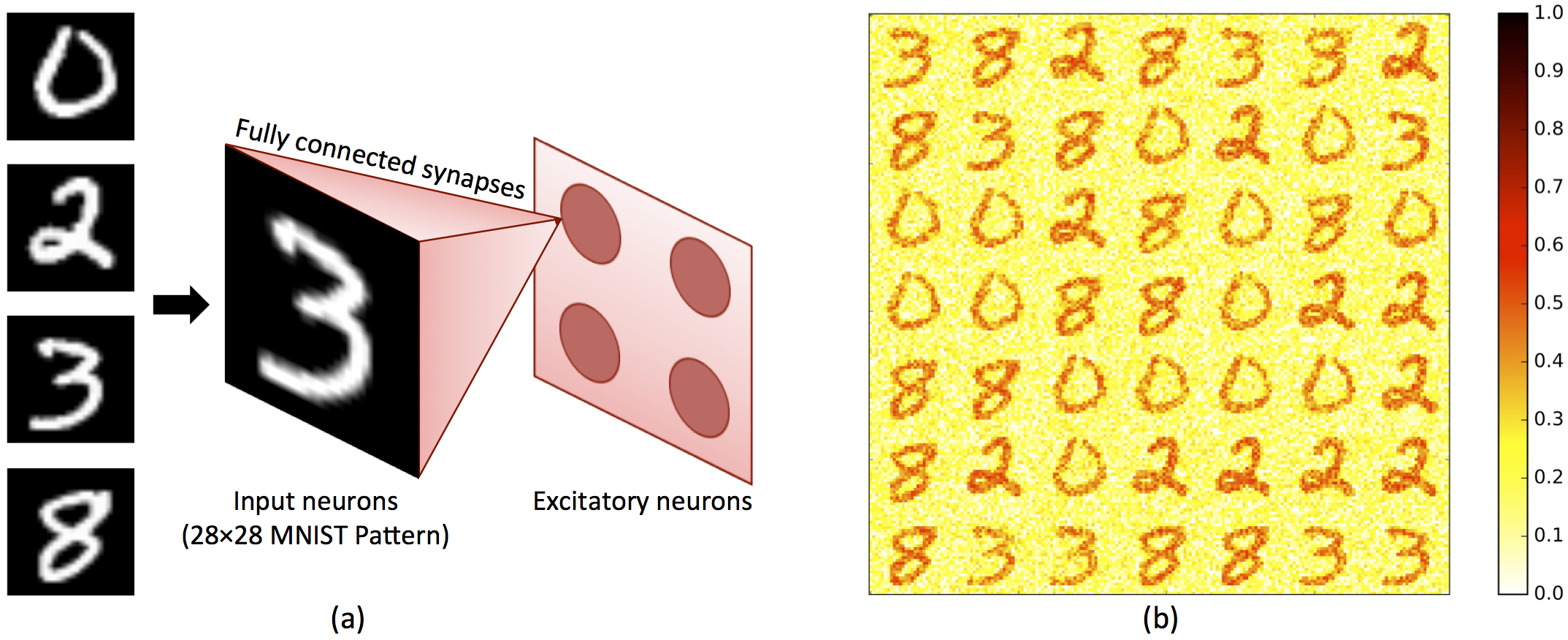}
\caption{\textbf{(a)} Fully-connected SNN topology for pattern recognition. The pixels in the image pattern to be recognized constitute the pre-neurons in the input layer that are fully-connected to the excitatory post-neurons. \textbf{(b)} Illustration of the synaptic learning mechanism in a fully-connected network of 50 excitatory neurons trained on a subset of digit patterns from the MNIST image set (28$\times$28 in resolution). The plot shows the final strength of the 784 excitatory synapses (rearranged in 28$\times$28 format) connecting each of the 50 post-neurons (organized in 7$\times$7 grid), which are adapted using the power-law weight-dependent plasticity model.}
\label{fig:FSNN}
\end{figure}

A hierarchical SNN topology (shown in Fig. \ref{fig:FSNN}(a)) consisting of a layer of input neurons fully-connected to the excitatory post-neurons has been widely explored for unsupervised pattern recognition \cite{diehl2015unsupervised}. The pixels in the image pattern to be recognized constitute the input neurons, each of which encodes the information contained in the corresponding image pixel as a Poisson-distributed spike train. The excitatory synaptic connections that are modulated using STDP based learning rules, consequently acquire a complete representation of the presented input patterns. Fig. \ref{fig:FSNN}(b) demonstrates the evolution of synaptic weights in a representative network of 50 excitatory neurons trained on a subset of handwritten digit patterns from the MNIST dataset \cite{lecun1998mnist}. It can be seen that the excitatory synapses that were randomly initialized in the beginning of the training phase, eventually encoded distinct digit patterns in the corresponding weights. We surmise that fully-connected SNNs learn a particular spatial arrangement of the input patterns, which leads to degradation in the recognition capability when evaluated on rotated variants of the training patterns. Furthermore, the fundamental limitation of learning in the fully connected topology is that it simply maps the given input image onto the synaptic weights and impedes the network from learning more generalized representations. This simplistic mapping scheme necessitates the network to be trained on larger number of training examples to learn varying representations of input patterns and yield competetive accuracy. To address this inadequacy, we explore a \textit{shared weight kernel feature extraction mechanism} with a Convolutional SNN topology. The goal of our convolutional learning scheme is to discover latent structures of inputs and learn underlying representations instead of learning the exact input pattern. This representation-learning perspective improves the synaptic learning efficiency and enables the network to learn from significantly lesser number of training examples.

\section*{\large\bf{Convolutional SNN Architecture}}
\begin{figure}[!t]
\centering
\includegraphics[width = 4.7in]{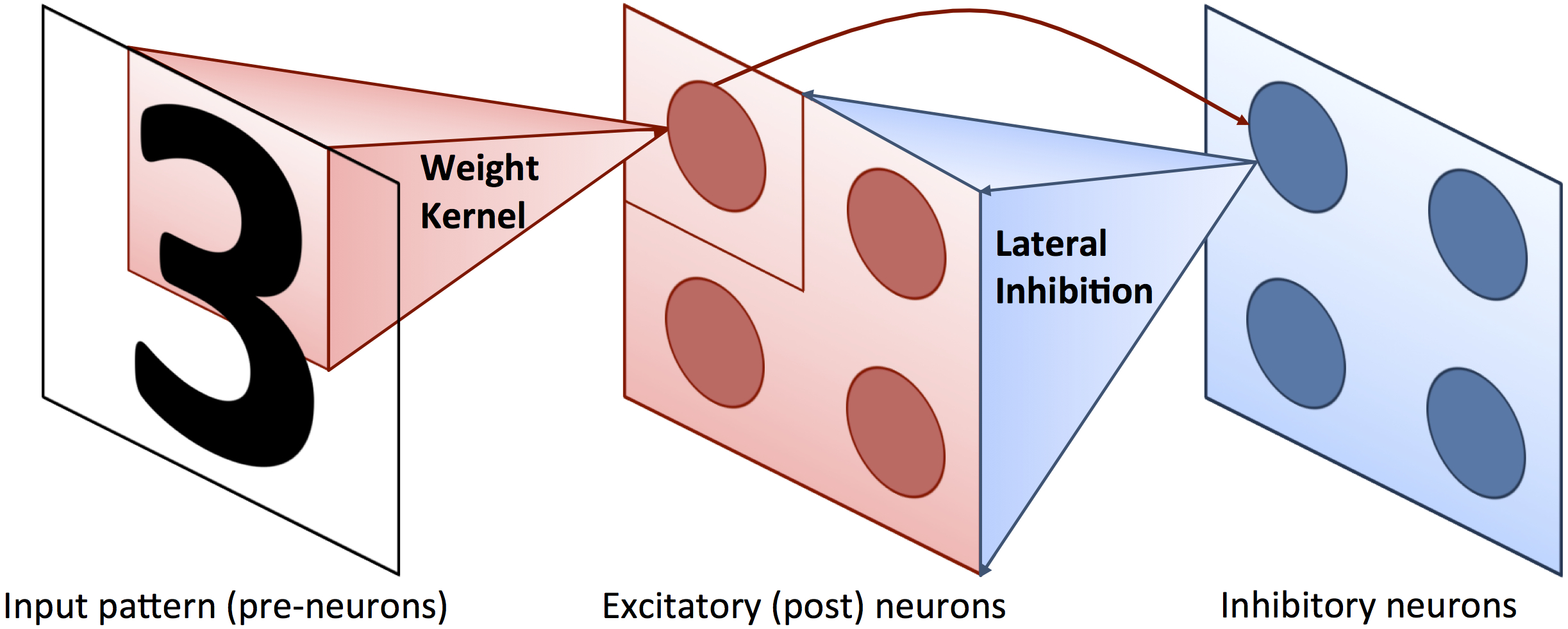}
\caption{Convolutional SNN topology for pattern recognition. The input neurons are connected to every post-neuron in the excitatory layer by a unique weight kernel that is shared among the pre-neurons. There are one-to-one connections between the excitatory neurons and the corresponding inhibitory neurons, each of which inhibits all the neurons in the excitatory layer except the one from which it received a forward connection.}
\label{fig:Conv_SNN}
\end{figure}

We present a Convolutional SNN architecture with sparsely connected input and excitatory layers to extract distinct features from the input patterns for efficient recognition. We use \textit{weight kernels} to achieve sparse synaptic connectivity between the input neurons and every excitatory post-neuron as illustrated in Fig. \ref{fig:Conv_SNN}. It is worth mentioning that the dimension of the weight kernel is smaller than the resolution of the input image pattern. We exploit the sharing of synaptic weights among the input neurons to extract salient features embedded in the presented patterns using the \textit{convolutional STDP learning methodology} that will be described in the following sub-section. It is important to note that each excitatory neuron has a unique synaptic kernel, which is trained to acquire attributes that characterize a specific class of input patterns. The excitatory neurons are further connected to the inhibitory neurons in a one-to-one manner. An excitatory neuronal spike eventually triggers the connected inhibitory neuron to fire, which subsequently inhibits the rest of the neurons in the excitatory layer. Lateral inhibition facilitates an excitatory neuron to spike exclusively for a distinct class of input patterns, which enables the corresponding weight kernel to learn the characteristic features.

\subsection*{\normalsize\bf{Convolutional STDP Learning Methodology}}

\begin{figure}[!t]
\centering
\includegraphics[width = 7.0in]{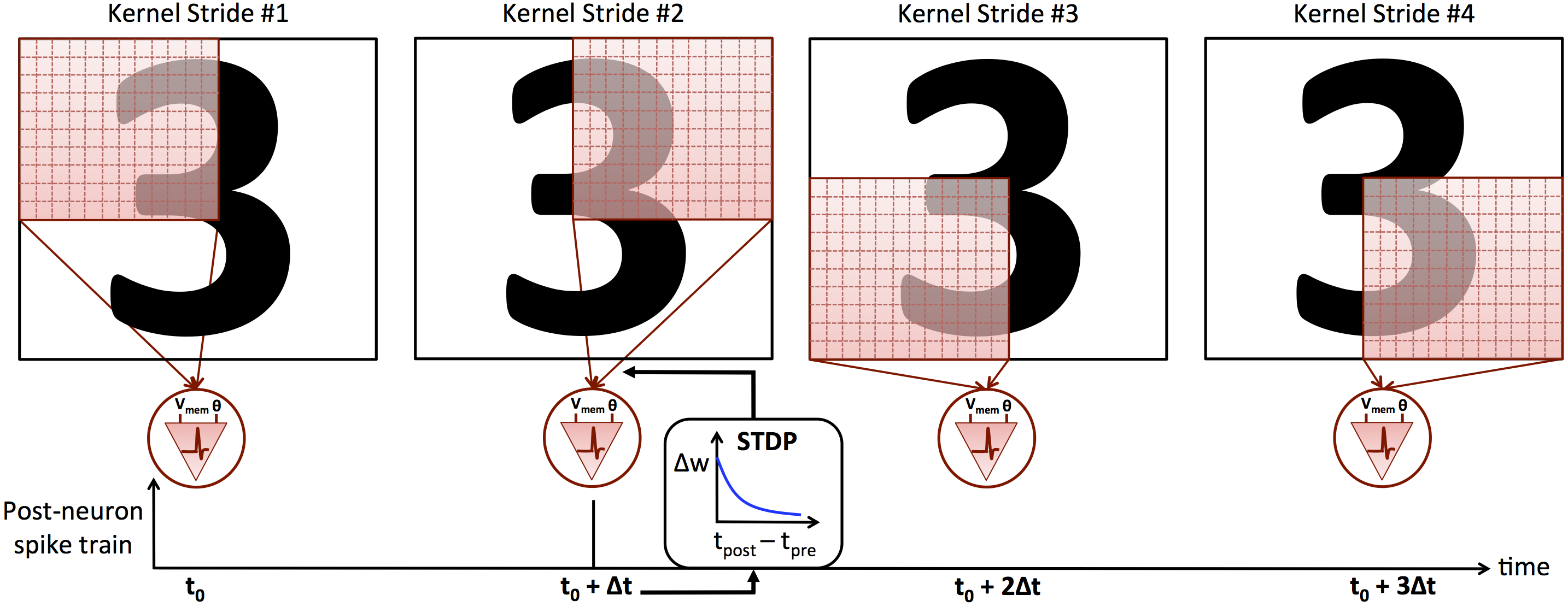}
\caption{Illustration of the convolutional STDP learning methodology for adapting the synaptic kernel connecting an input pattern to a specific excitatory post-neuron. At every time-step, the pre-neurons belonging to a distinct region in the input image are modulated by the kernel weights to generate a resultant current into the post-neuron. This increases the neuronal membrane potential ($V_{mem}$), which triggers an output spike if the potential exceeds a definite threshold ($\theta$). The kernel weights are modified at the instant of a post-neuronal spike ($t_{0}+\Delta t$ in this example) based on the temporal correlation with the corresponding pre-spikes as stipulated by the power-law weight-dependent plasticity model. The kernel is moved over (convolved with) the entire image by a pre-specified number of strides (4 in this example) over multiple time-steps. This operation is repeated across the entire time duration for which the input pattern is presented.}
\label{fig:Conv_STDP}
\end{figure}

In this sub-section, we detail the proposed convolutional STDP learning methodology for self-learning the synaptic weight kernels used for input feature extraction. We begin by describing the learning methodology for a representative weight kernel connecting the input neurons to a single excitatory neuron, and subsequently demonstrate its applicability in training different kernels in a network of such spiking neurons.

Consider a randomly initialized weight kernel connecting a segment of the presented input pattern to an excitatory post-neuron as shown in Fig. \ref{fig:Conv_STDP}. Every pixel in the image pattern constitutes an input pre-neuron, whose firing rate is proportional to the corresponding pixel intensity. At any given time, the input pre-neuronal spikes are modulated by the kernel weights to produce a resultant current into the post-neuron, which increases the neuronal membrane potential. The post-neuron fires a spike once the potential crosses a definite threshold. In the absence of a post-neuronal spiking event, the kernel is simply moved over the image by a pre-specified number of strides. However, in the event of a post-neuronal spike (time instant $t_{0}+\Delta t$ in Fig. \ref{fig:Conv_STDP}), the kernel weights are modified based on the temporal correlation between the post-spike and the corresponding pre-spikes as specified by the power-law weight-dependent STDP model, in the previous section. The updated weight kernel influences the neuronal spiking dynamics at the following time instant. We note that the kernel is moved sequentially over the entire image by a fixed number of horizontal and vertical strides across multiple time-steps. In other words, the kernel is \textit{convolved} with the input pattern over several instants of time depending on the number of strides. The convolution operation is performed repeatedly until the end of the training duration.

In the presented learning methodology, it is important to note that the synaptic kernel is connected to a distinct region in the input image at every time-step. Hence, updating the kernel at various instants of a post-neuronal spike enables it to encode features located in different image segments. The sharing of synaptic weights among the input neurons effectively causes the kernel to learn an overlapped representation of the characteristic features in the presented pattern.

\begin{figure}[!t]
\centering
\includegraphics[width = 5.0in]{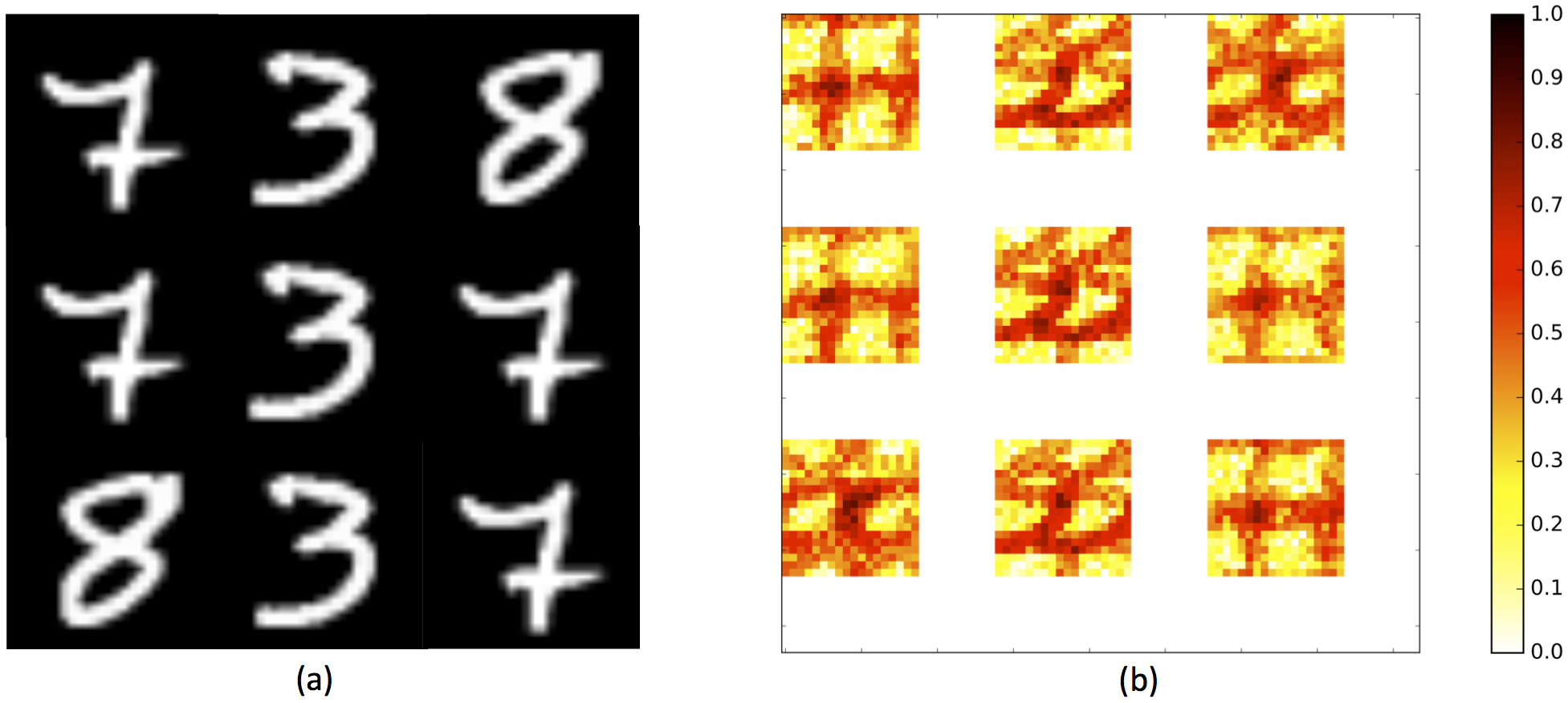}
\caption{\textbf{(a)} MNIST digit patterns (arranged in 3$\times$3 matrix) used to train a Convolutional SNN of 10 excitatory neurons. \textbf{(b)} Synaptic kernels (18$\times$18 in size) associated with each of the 10 excitatory neurons (organized in 3$\times$3 grid) trained using the convolutional STDP learning methodology on the corresponding input digit patterns.}
\label{fig:Conv_MNIST_Kernels1}
\end{figure}

Finally, we demonstrate the applicability of the proposed learning mechanism in adapting the synaptic kernels associated with every excitatory neuron in a Convolutional SNN. At any instant of time, the various kernels are convolved with pre-neurons belonging to a specific region in the input image to produce a resultant current into the respective post-neurons. The SNN topology is additionally equipped with lateral inhibitory connections, which causes a distinct excitatory neuron to spike at a higher rate and consequently learn the presented image pattern. Lateral inhibition, in effect, differentiates the receptive field of every neuron in the excitatory layer and helps achieve competitive learning. At the end of the training phase, the synaptic kernel associated with each neuron learns salient features embedded in a unique class of input patterns. This is illustrated in Fig. \ref{fig:Conv_MNIST_Kernels1}, which shows the kernels learned by a Convolutional SNN of 10 excitatory neurons trained on a subcategory of digit patterns from the MNIST dataset \cite{lecun1998mnist}. We use the kernel representation of digit patterns `$3$', `$7$', and `$8$' to intuitively highlight the feature extraction capability of the presented convolutional learning methodology. It can be seen from Fig. \ref{fig:Conv_MNIST_Kernels1}(b) that the kernel associated with `$3$' learned distinct semi-closed loops, while that linked to `$8$' acquired a couple of circular loops. The kernel related to `$7$' encoded the individual horizontal and vertical lines in the digit pattern. We show in the experimental results section that the characteristic features thus learned can be used to efficiently recognize the various input patterns across multiple datasets.

\subsection*{\normalsize\bf{Convolutional versus Fully-connected Topology}}
We now highlight the essential differences and consequent benefits of the proposed Convolutional SNN compared to a typical fully-connected architecture (shown in Fig. \ref{fig:FSNN}). First, the Convolutional SNN provides sparser synaptic connectivity by virtue of using smaller weight kernels between the input and excitatory neurons as opposed to a fully-connected SNN possessing all-to-all connections. This lowers the synaptic storage requirement and accordingly the area footprint of synaptic memory in custom hardware realizations of large-scale SNNs. Furthermore, the presented topology offers intrinsic energy efficiency that primarily stems from reduction in the number of synapses that need to be accessed and subsequently programmed based on STDP in the event of a post-neuronal spike. Second, the Convolutional SNN demonstrates improved synaptic learning capability. This can fundamentally be attributed to the ability of weight kernels to encode significant features characterizing various input image patterns, which is in contrast to a fully-connected SNN that learns particular spatial representations. As a result, the Convolutional SNN can potentially achieve comparable classification accuracy to that of a fully-connected network using fewer number of training examples as will be shown in the experimental results section. In other words, given a limited training dataset, the Convolutional SNN provides relatively higher classification accuracy.

Last, the Convolutional SNN exhibits \textit{rotational invariance} in pattern recognition tasks, i.e., the network is capable of recognizing rotated versions of image patterns used during the training phase. The convolution of a trained synaptic kernel with an input pattern enables the connected post-neuron to detect similar characteristic features present in the rotated image pattern. On the other hand, a fully-connected SNN is inherently sensitive to image orientation, and therefore needs to be trained on the rotated patterns for accurate classification. It is important to note that the recognition efficiency of the Convolutional SNN strongly depends both on the kernel size and the number of strides needed to slide over the entire image. We comprehensively analyze the trade-offs between classification accuracy and kernel configuration in the experimental results section.

\section*{\large\bf{Simulation Framework}}
The presented Convolutional SNN topology consisting of the input, excitatory, and inhibitory layers is implemented in an open-source Python-based simulation framework, BRIAN \cite{goodman2008brian}, for recognizing patterns from MNIST \cite{lecun1998mnist}, face detection \cite{YUV_FACE}, and Caltech \cite{fei2007learning} datasets. The leaky integrate-and-fire dynamics of the excitatory and inhibitory neurons are modeled using differential equations with parameters essentially adopted from \cite{jug2012competition}. The simulator is further augmented to effect sparse connectivity between the input and excitatory neurons using the proposed synaptic kernels, each of which is convolved with a given input pattern over multiple simulation time-steps. The individual weights constituting a kernel are modified at the instants of every spike fired by the connected post-neuron based on the temporal correlation with the corresponding pre-neuronal spikes. The degree of correlation between a pair of pre- and post-neuronal spikes is obtained by generating an exponentially decaying \textit{pre-trace} that integrates the input spikes, and sampling it at the instant of a post-spike as described in \cite{diehl2015unsupervised}.

The convolutional STDP learning methodology effectively causes the various randomly initialized synaptic kernels to encode characteristic features pertaining to distinct classes of input patterns. For multiclass recognition tasks, for instance, MNIST digit recognition, every excitatory post-neuron is tagged as having learned a particular class for which it spiked the most during the training phase. The recognition performance of the trained network is subsequently estimated on a separate testing dataset. Each test image is predicted to belong to the class represented by the group of neurons with the highest average spike-count throughout the simulation period. The classification is accurate if the actual class of a test image matches with that predicted by the network of spiking neurons.

We now detail the training and testing procedure for face detection, which is a binary classification task consisting of facial and background images. The Convolutional SNN is trained exclusively on different facial patterns. During the evaluation phase, an input image is predicted to depict a face if the average spike-count of the excitatory neurons over the simulation interval exceeds a pre-determined rate of firing. On the contrary, if the spiking activity is insufficient as determined by the threshold firing rate, the input is deemed a background image pattern. The classification accuracy, which is defined as the ratio of the number of images correctly classified by the trained network to the total number of test images, is used to validate the effectiveness of the convolutional STDP learning methodology on the MNIST and face detection datasets. We use few sample patterns from the Caltech dataset to demonstrate the rotational-invariant recognition capability of the Convolutional SNN.

\section*{\large\bf{Experimental Results}}
\subsection*{\normalsize\bf{MNIST Digit Recognition}}
\begin{figure}[!t]
\centering
\includegraphics[width = 5.5in]{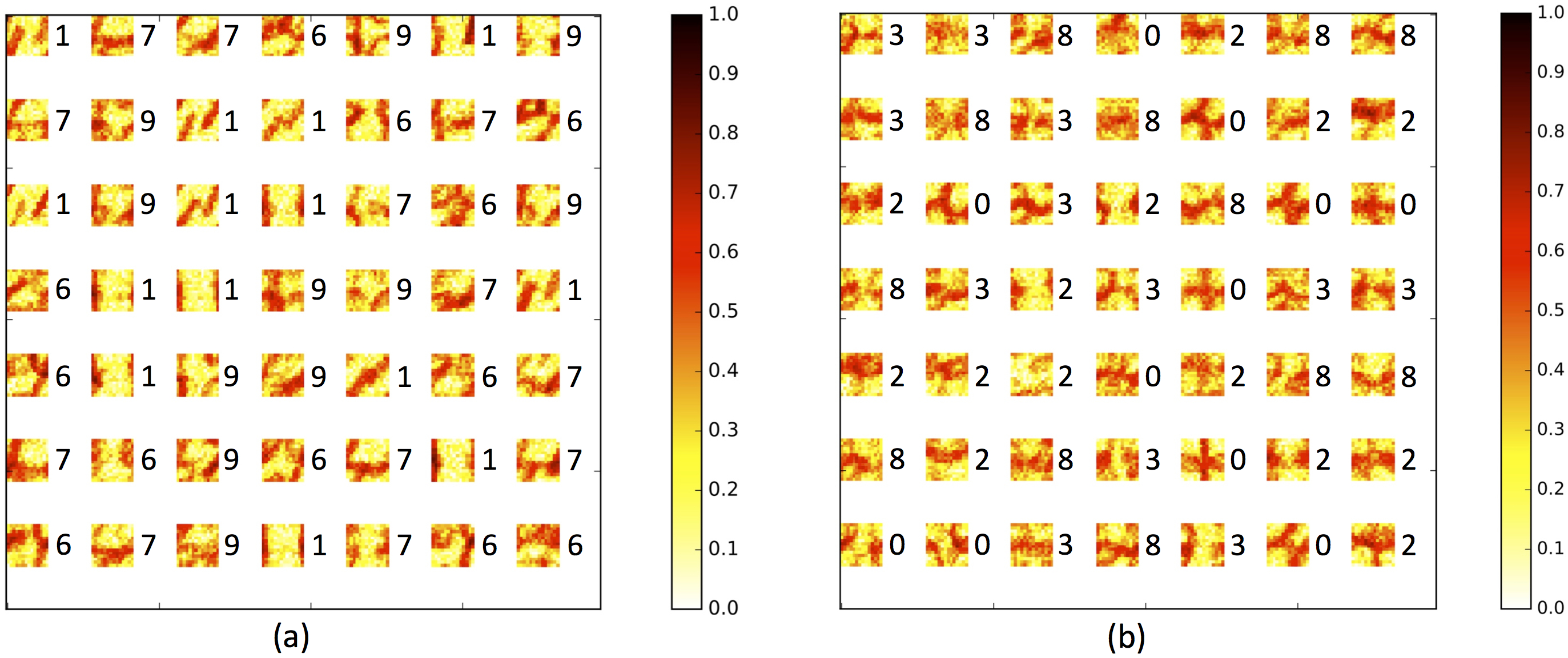}
\caption{\textbf{(a)} Synaptic kernels (14$\times$14 in size) and the associated output classes (digits) learned by a network of 50 excitatory neurons (organized in 7$\times$7 grid) trained on the subset \{`$1$', `$6$', `$7$', `$9$'\} using 40 examples from each category. \textbf{(b)} Synaptic kernels (14$\times$14 in size) and the corresponding output classes (digits) learned by a similarly sized network trained on the subset \{`$0$', `$2$', `$3$', `$8$'\} using 40 examples from each category.}
\label{fig:Conv_MNIST_Kernels2}
\end{figure}

The MNIST image recognition dataset \cite{lecun1998mnist} consists of 60000 training and 10000 testing examples of different handwritten digits, where each digit pattern is 28$\times$28 in dimension. Initially, we use a smaller network trained on a subset of output classes (digits) to illustrate the feature extraction capability of the presented Convolutional SNN architecture. A smaller albeit sufficiently complex training subset is primarily chosen to be able to comprehensively analyze various kernel sizes and strides at a reasonable simulation speed for identifying the optimal configuration. We subsequently compare the classification accuracy of the optimized Convolutional SNN topology with that of fully-connected network of similar size to show its improved efficiency in pattern recognition applications. Finally, we train the Convolutional SNN on all the output classes (`$0$'--`$9$') for demonstrating its applicability in larger multiclass recognition tasks.

In our first experiment, we trained a Convolutional SNN of 50 excitatory neurons using 40 unique patterns each of digits `$1$', `$6$', `$7$', and `$9$'. The selected training subset consists of patterns sharing common features (`$1$', `$7$', `$9$') along with those that are complementary (`$9$', `$6$'). It can be seen from Fig. \ref{fig:Conv_MNIST_Kernels2}(a) that the weight kernel associated with each neuron encoded features corresponding to a specific output class (printed alongside each kernel). Our simulations further indicated that the impact of lateral inhibition needed to be high for efficient synaptic learning. We note that a stronger inhibitory strength causes a distinct excitatory neuron to spike at a higher rate and predominantly learn a given input pattern while inhibiting the rest of the network. This enables the convolutional STDP learning mechanism to precisely adapt the weight kernels connecting every individual neuron in the excitatory layer. The trained network offered a classification accuracy of 92.45\% on all the 4000 images of `$1$', `$6$', `$7$', and `$9$' in the MNIST testing dataset, which illustrates the competence of the Convolutional SNN to distinguish between patterns containing shared features. A similar analysis on a network of 50 neurons using a more complex training subset consisting of the digits `$0$', `$2$', `$3$', and `$8$' (Fig. \ref{fig:Conv_MNIST_Kernels2}(b)) provided a classification accuracy of 83.25\%.

\begin{figure}[!t]
\centering
\includegraphics[width = 7.0in]{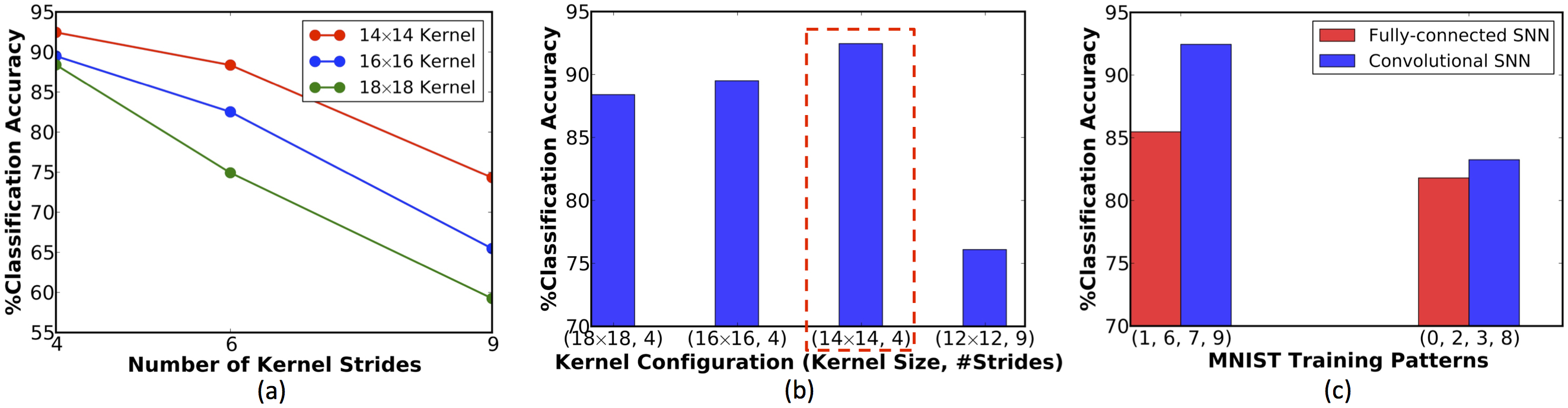}
\caption{\textbf{(a)} Classification accuracy of a Convolutional SNN of 50 excitatory neurons trained using 40 unique patterns each of digits `$1$', `$6$', `$7$', and `$9$' versus the number of kernel strides. \textbf{(b)} Classification accuracy of the network of 50 neurons for different kernel configurations including its size and the optimal number of strides. \textbf{(c)} Classification accuracy of the optimized Convolutional SNN topology (14$\times$14 kernel with 4 strides) and the baseline fully-connected SNN individually trained on the subsets \{`$1$', `$6$', `$7$', `$9$'\} and \{`$0$', `$2$', `$3$', `$8$'\} using 40 distinct examples from each class of digits.}
\label{fig:Conv_MNIST_Results1}
\end{figure}

\begin{figure}[!t]
\centering
\includegraphics[width = 6.0in]{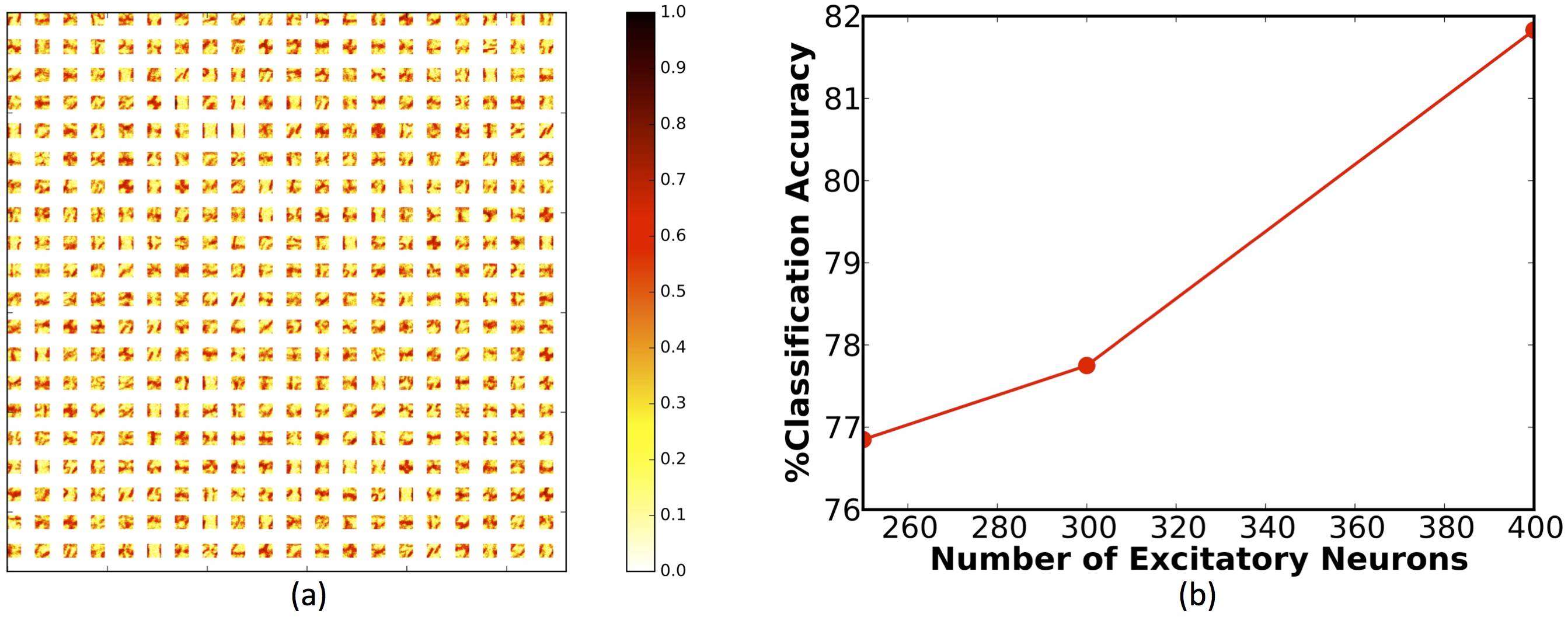}
\caption{\textbf{(a)} Synaptic kernels (14$\times$14 in size) learned by a network of 400 excitatory neurons (organized in 20$\times$20 grid) trained using 80 examples from each class of digit patterns from `$0$' to `$9$'. \textbf{(b)} Classification accuracy of the optimized Convolutional SNN topology (14$\times$14 kernel with 4 strides) trained on all the digits (`$0$'--`$9$') using 80 examples from each output class against the number of excitatory neurons.}
\label{fig:Conv_MNIST_Results2}
\end{figure}

Next, we perform a rigorous analysis on a Convolutional SNN of 50 excitatory neurons using the training subset comprising digits `$1$', `$6$', `$7$', and `$9$' to ascertain the optimal kernel size and the number of horizontal ($h$) and vertical ($v$) strides required to cover the entire image. We investigated strides of 4 (2$h$, 2$v$), 6 (3$h$, 2$v$), and 9 (3$h$, 3$v$) for three different sizes of the synaptic kernel (18$\times$18, 16$\times$16, and 14$\times$14). Our results (Fig. \ref{fig:Conv_MNIST_Results1}(a)) indicate a drop in the classification accuracy as the number of strides is increased. The highest classification accuracy is obtained for a stride configuration of 4 for all the kernel sizes under investigation. In other words, we find that it is desirable to have minimal number of strides needed to cover the image pattern for a given size of the synaptic kernel. For instance, a 14$\times$14 kernel requires a minimum of 4 strides while a 12$\times$12 kernel necessitates 9 strides over the MNIST image (28$\times$28 in dimension). We now explore kernels of varying sizes, each of which is convolved with an input pattern using the optimal number of strides. We note from Fig. \ref{fig:Conv_MNIST_Results1}(b) that the classification accuracy improves as the kernel size is reduced from 18$\times$18 to 14$\times$14, and degrades significantly ($>$15\%) for sizes of 12$\times$12 and beyond. We surmise that the favorable configuration for a synaptic kernel including its size and the number of strides is the one that results in minimal kernel overlap across adjacent strides over an image pattern. We observed in our simulations that an increase in kernel overlap during the striding operation leads to excessive weight updates, which deteriorate the learned features. For MNIST digit recognition, a 14$\times$14 kernel with 4 strides over distinct non-overlapping image regions offers the best classification accuracy. It is worth mentioning that decreasing the kernel size below 12$\times$12 failed to elicit sufficient number of spikes over the simulation interval, thereby proving detrimental to synaptic learning.

The optimized Convolutional SNN topology trained using 40 unique patterns on the subsets \{`$1$', `$6$', `$7$', `$9$'\} and \{`$0$', `$2$', `$3$', `$8$'\} respectively provided 7\% and 1.45\% increase in classification accuracy over a fully-connected network of equivalent size as shown in Fig. \ref{fig:Conv_MNIST_Results1}(c). This highlights the superior recognition capability of the Convolutional SNN given a limited number of training examples. The enhancement in classification accuracy is achieved with 4$\times$ sparser synaptic connectivity between the input and excitatory neurons, where the sparsity metric is defined as the ratio of number of synaptic connections per neuron in a fully-connected network to the kernel size in the Convolutional SNN topology.

Finally, we trained a larger network on all the digits from `$0$' to `$9$' using the optimal kernel configuration previously determined (14$\times$14 kernel with 4 (2$h$, 2$v$) strides) to verify its efficacy in complex multiclass recognition tasks. Fig. \ref{fig:Conv_MNIST_Results2}(a) shows the distinct synaptic kernels learned by a network of 400 excitatory neurons trained with 80 examples from every class of digit patterns. We obtained a classification accuracy of 81.8\% on the entire MNIST testing dataset while merely using 800 training examples out of a total of 60000. We note that a similarly sized fully-connected SNN needs to be trained on approximately 6000 training patterns to achieve comparable accuracy. Thus, the ability of our Convolutional SNN to attain nominal classification accuracy using fewer training images is a testament to its improved feature extraction capacity rendered possible by the presented learning methodology. The added sparsity in synaptic connectivity can be exploited to achieve energy efficiency in custom hardware implementations. Lastly, we note that the classification accuracy can be improved further by increasing the number of excitatory neurons as depicted in Fig. \ref{fig:Conv_MNIST_Results2}(b).

\subsection*{\normalsize\bf{Face Detection}}
\begin{figure}[!t]
\centering
\includegraphics[width = 6.5in]{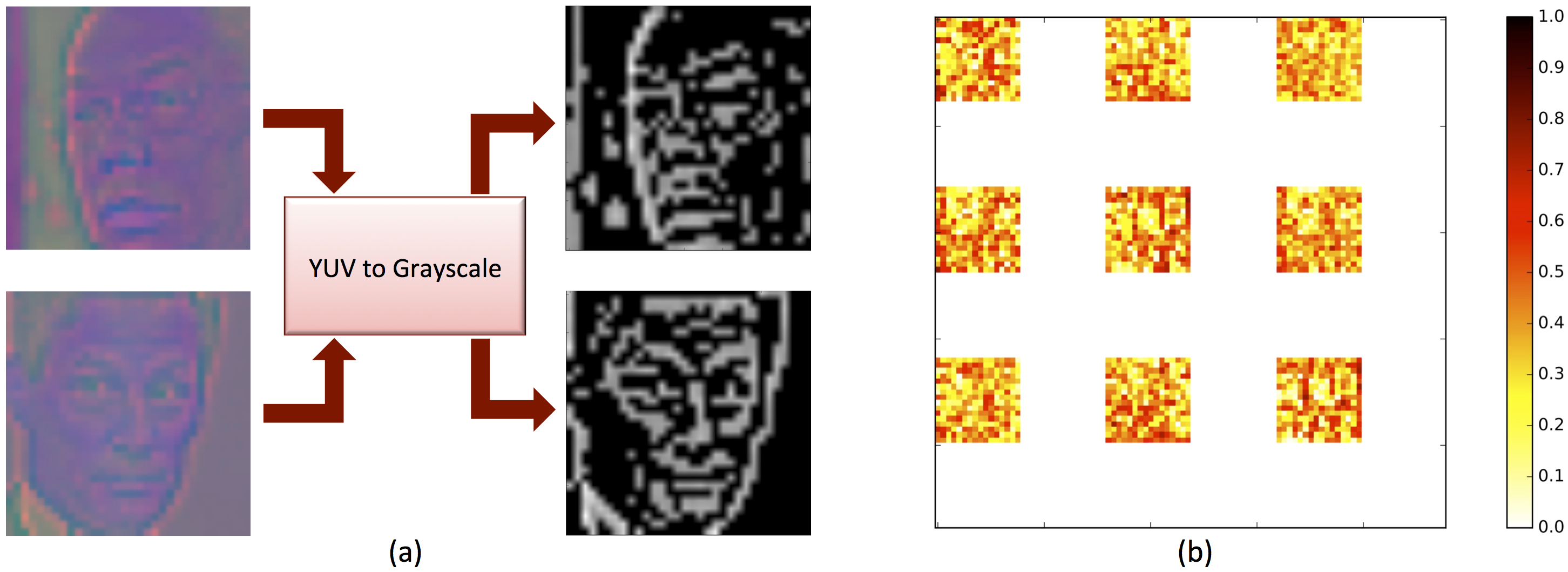}
\caption{\textbf{(a)} Input pre-processing procedure for the face detection task, where the image samples (32$\times$32 in resolution) are converted from $YUV$ colorspace to grayscale format. \textbf{(b)} Synaptic kernels (16$\times$16 in size) learned by a Convolutional SNN of 10 excitatory neurons trained using 10 distinct facial patterns.}
\label{fig:Conv_FaceYUV_Results}
\end{figure}

The face detection dataset \cite{YUV_FACE} consists of 13000 facial and 28000 random background images. Each 32$\times$32 image pattern is stored in $YUV$ colorspace, where $Y$ denotes the luminance (brightness) component while $U$ and $V$ correspond to the chrominance (color) components. We pre-process the $YUV$ image as explained below for compatibility with the input encoding scheme adopted in this work, where every pixel in a single channel grayscale image is converted to a train of spikes feeding the excitatory neurons. First, we extract the luminance ($Y$) component from the original image. Next, we apply a pixel thresholding technique, where only those pixels with intensity greater than a definite threshold are retained while the remaining pixels are blackened (intensity forced to 0). The pre-processing step effectively transforms the image patterns from $YUV$ colorspace to grayscale format while accentuating the significant input features as illustrated in Fig. \ref{fig:Conv_FaceYUV_Results}(a).

We subsequently trained a Convolutional SNN of 10 excitatory neurons on 10 distinct facial images using 16$\times$16 kernels striding 4 times over each input pattern. Fig. \ref{fig:Conv_FaceYUV_Results}(b) shows the features learned by the weight kernels at the end of the training phase. The efficiency of synaptic learning is evaluated on a testing dataset consisting of 500 samples from each of the facial and background classes. An input image is predicted to represent a facial pattern only if 50\% of the excitatory neurons fire at least once during the simulation period. The network trained solely on 10 facial examples yielded a classification accuracy of 79.3\% on 1000 test images with 4$\times$ sparsity in the synaptic connections between input and excitatory layer. This reinforces the capacity of the Convolutional SNN to achieve competitive classification accuracy using fewer training patterns while providing sparse synaptic connectivity.

\subsection*{\normalsize\bf{Caltech Image Recognition}}
\begin{figure}[!t]
\centering
\includegraphics[width = 6.8in]{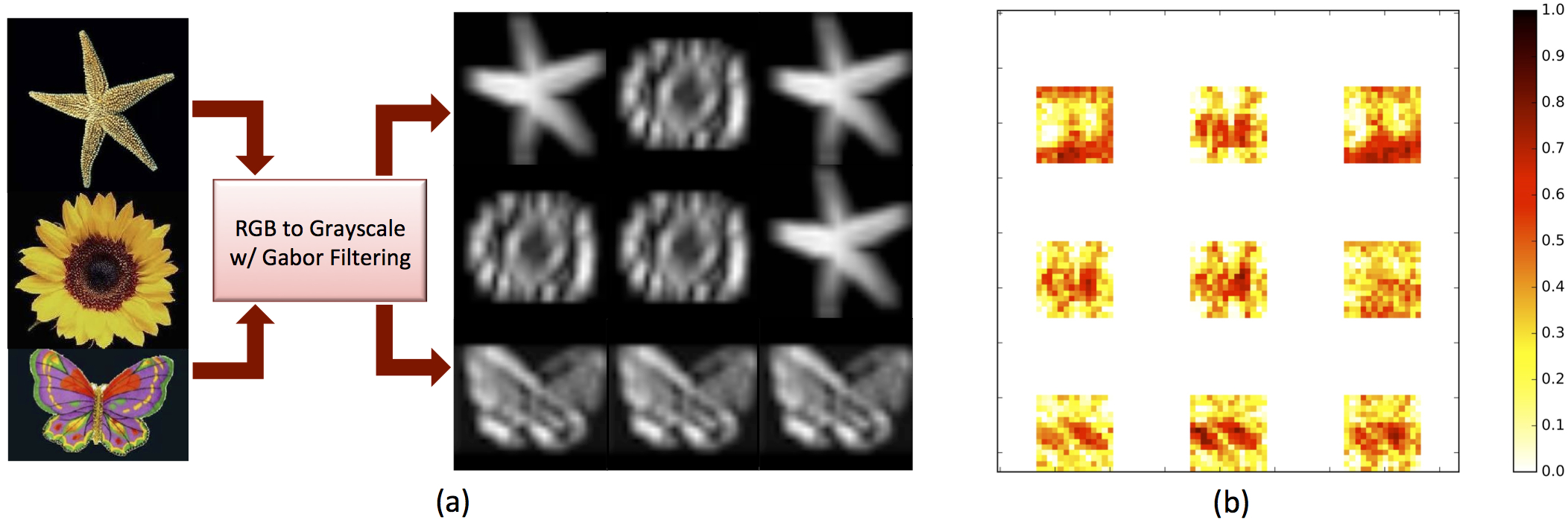}
\caption{\textbf{(a)} Input pre-processing procedure for the Caltech image recognition task, where the original high resolution RGB images are filtered using Gabor kernels to produce grayscale images of reduced dimension (32$\times$32). \textbf{(b)} Synaptic kernels (16$\times$16) learned by a network of 10 excitatory neurons trained using the corresponding pre-processed image patterns from the Caltech dataset.}
\label{fig:Conv_Caltech_Results}
\end{figure}

We use few randomly selected images from the Caltech dataset \cite{fei2007learning} to demonstrate the rotational invariant recognition capability offered by the proposed Convolutional SNN topology. This is accomplished by training the network on the original image patterns and evaluating its classification performance on a testing set containing rotated versions of different training examples. The Caltech dataset is composed of color images stored in high resolution RGB colorspace (for instance, 230$\times$200$\times$3). We pre-processed the RGB image by applying Gabor filtering \cite{jain1997object} technique to convert it to a single channel grayscale image of lower dimension (32$\times$32), which retains the salient features as illustrated in Fig. \ref{fig:Conv_Caltech_Results}(a).

For this task, we trained a Convolutional SNN of 10 excitatory neurons using 3 distinct pre-processed Caltech image patterns shown in Fig. 10 (a). Initially, we selected 16$\times$16 synaptic kernels with each striding 4 times over an input pattern. The extent of rotational invariance offered by the trained network is evaluated on original patterns rotated individually by 45$^\circ$ and 90$^\circ$. Our simulations indicated that the stride configuration of 4 could accurately recognize image patterns rotated by 90$^\circ$ but proved ineffective for those rotated by 45$^\circ$. Intuitively, convolving a kernel using 4 strides over an input pattern, which involves 2 strides respectively in the horizontal and vertical dimensions matches the encoded features primarily in 4 image quadrants. This in turn enables the network to recognize patterns rotated by angles in close proximity to 90$^\circ$ or its multiples. For the network to correctly recognize patterns rotated by other intermediate angles, we hypothesize that the number of strides ought to be increased. We trained the kernels (Fig. \ref{fig:Conv_Caltech_Results}(b)) using 6 strides for each convolution operation. This kernel configuration could recognize patterns rotated both by 45$^\circ$ and 90$^\circ$ with an accuracy of 87.5\%. Fully-connected SNNs, on the other hand, need to be separately trained on the rotated patterns for accurate recognition since they learn definite spatial representations.

\subsection*{\normalsize\bf{Out-of-Set Generalization with Convolutional STDP}}
A fundamental drawback of machine learning models is that the system has to encounter all possible classes during training (that it will face later in the testing phase) in order to yield competitive accuracy on the testing dataset. This issue is generally termed as out-of-set generalization. While learning with fewer training examples and rotational invariance obtained from our convolutional learning scheme are a significant advancement to prior research on SNNs, the generic representation learning further equips the network to perform out-of-set generalization \cite{cox2014neural}.

To illustrate this, we trained a simple Convolutional SNN of 10 excitatory neurons using 10 unique patterns each of digits (`6', `7') from the MNIST dataset. During testing, we showed two completely different classes of digits (`9', `1') and measured the accuracy of the network on 2000 test patterns from these classes that the network has never encountered. The network accuracy was approximately 94.4\%. Further, our simulation results showed that the neurons that learnt digit `6' (`7') spiked the most for `9' (`1') during the testing phase. This implies that the network was able to associate the underlying representations learnt in the weight kernels for `6' (`7') with `9' (`1'). This further corroborates the efficacy of the features extracted with the weight kernels from our Convolutional STDP scheme. A similar experiment on a 10 excitatory neuron SNN with training on digits (`2', `8') and testing on (`0', `3') yielded 74.6\% accuracy. We observed that the neurons that learnt digit `8' spiked for a considerable number of test patterns of `3'. However, a large fraction of `0' in the testing set was not able to initiate an optimal spike response in the SNN that resulted in a decline in accuracy. This shows that the weight kernels learnt could not be associated with the digit `0'.  Thus, the SNN’s ability to perform out-of-set generalization is also limited based on the type of input patterns presented during testing. However, we would like to emphasize that for judiciously selected patterns, the Convolutional SNN can perform out-of-set generalization and yield competitive accuracy for a wide-range of input patterns. This primary quality cannot be observed with typical fully-connected SNN architectures since the weights learnt are more representative of the entire image rather than underlying structures. In fact, artificial deep learning models also cannot perform out-of-set generalization without undergoing some retraining at the later layers of the hierarchy \cite{scheirer2014probability}. 

\section*{\large\bf{Conclusion}}
SNNs constitute a powerful computing paradigm as they offer a promising solution to approaching brain’s efficient processing capability for cognition. Most unsupervised SNN architectures explored for recognition applications use a fully-connected topology that makes them inefficient with regard to synaptic storage. Moreover, the fully-connected architecture fails to learn characteristic representations underlying the image patterns. As a result, they require training on a large fraction of images that adds to the inefficiency of the learning model. In this work, we proposed a Convolutional SNN architecture that uses shared weight kernels to learn the representative features from visual inputs in an unsupervised manner with STDP. The proposed Convolutional SNN offers lower storage requirement in addition to improved learning with lesser number of training patterns. Our results on three benchmark applications demonstrate that Convolutional SNN consistently outperforms conventional SNN architectures in terms of output quality despite being trained on $\sim$10 times lesser number of examples than the latter. We would like to emphasize that the convolutional STDP learning enables the network to recognize objects in spite of different variations thereby improving the robustness of the learning model. Finally, the Convolutional SNN with biologically plausible neuronal functions and learning mechanism performs preliminary out-of-set generalization further underpinning the effectiveness of our methodology.

\section*{\large\bf{Acknowledgments}}
The work was supported in part by, Center for Spintronic Materials, Interfaces, and Novel Architectures (C-SPIN), a MARCO and DARPA sponsored StarNet center, by the Semiconductor Research Corporation, the National Science Foundation, Intel Corporation, and the DoD Vannevar Bush Fellowship.

\section*{\large\bf{Author Contributions}}
P.P. and K.R. conceived the study. G.S. designed the framework and performed the simulations. P.P. assisted in performing the simulations. All authors helped with the writing of the manuscript, and discussing the results.
\section*{\large\bf{Competing Financial Interests}}
The authors declare no competing financial interests.





%

\end{document}